%% file: ReActor.tex
\documentclass[acmtog]{acmart}
\acmSubmissionID{1360}

\usepackage[ruled]{algorithm2e} % For algorithms

\SetAlFnt{\small}
\SetAlCapFnt{\small}
\SetAlCapNameFnt{\small}
\SetAlCapHSkip{0pt}

\usepackage{graphicx}
\usepackage{siunitx}
\usepackage{float}
\usepackage{multirow}

\usepackage{amsmath}
\usepackage{mathtools}
\usepackage{siunitx}
\usepackage{bm}
\newcommand{\OURMETHOD}{\textsc{ReActor}}

\definecolor{rebuttal}{RGB}{0, 0, 0}

\newcommand{\figref}[1]{Fig.~\ref{#1}}
\newcommand{\secref}[1]{Sec.~\ref{#1}}

\newcommand{\eqnref}[1]{Eq.~\eqref{#1}}
\newcommand{\tabref}[1]{Tab.~\ref{#1}}

\newcommand{\vect}{\mathbf}	% define \vect -> bold
\definecolor{paramPurple}{RGB}{148, 33, 146} % Purple for optimized variables
\definecolor{nomBlue}{RGB}{0, 118, 186} % Blue for nominal parts of the mapping

\begin{document}
% Title portion

\title{ReActor: Reinforcement Learning for Physics-Aware Motion Retargeting}

\author{David M{\"u}ller}
\email{david.cao@disneyresearch.com}
\orcid{0009-0001-6591-8803}
\affiliation{%
  \institution{Disney Research}
  \country{Switzerland}
}

% DO NOT ENTER AUTHOR INFORMATION FOR ANONYMOUS TECHNICAL PAPER SUBMISSIONS TO SIGGRAPH!
\author{Agon Serifi}
\email{agon.serifi@disneyresearch.com}
\orcid{0000-0003-4439-0023}
\affiliation{%
  \institution{Disney Research}
  \country{Switzerland}
}

\author{Sammy Christen}
\email{sammy.christen@disneyresearch.com}
\orcid{0000-0002-3511-8565}
\affiliation{%
  \institution{Disney Research}
  \country{Switzerland}
}

\author{Ruben Grandia}
\email{ruben.grandia@disneyresearch.com}
\orcid{0000-0002-8971-6843}
\affiliation{%
  \institution{Disney Research}
  \country{Switzerland}
}

\author{Espen Knoop}
\email{espen.knoop@disneyresearch.com}
\orcid{0000-0002-7440-5655}
\affiliation{%
  \institution{Disney Research}
  \country{Switzerland}
}

\author{Moritz B{\"a}cher}
\email{moritz.baecher@disneyresearch.com}
\orcid{0000-0002-1952-1266}
\affiliation{%
  \institution{Disney Research}
  \country{Switzerland}
}

\begin{abstract}
Retargeting human kinematic reference motion onto a robot's morphology remains a formidable challenge. Existing methods often produce physical inconsistencies, such as foot sliding, self-collisions, or dynamically infeasible motions, which hinder downstream imitation learning. We propose a bilevel optimization framework that jointly adapts reference motions to a robot’s morphology while training a tracking policy using reinforcement learning. To make the optimization tractable, we derive an approximate gradient for the upper-level loss. Our framework requires only a sparse set of semantic rigid-body correspondences and eliminates the need for manual tuning by identifying optimal values for a parameterization expressive enough to preserve characteristic motion across different embodiments. Moreover, by integrating retargeting directly with physics simulation, we produce physically plausible motions that facilitate robust imitation learning. We validate our method in simulation and on hardware, demonstrating challenging motions for morphologies that differ significantly from a human, including retargeting onto a quadruped.
\end{abstract}

% Copyright Siggraph
\setcopyright{cc}
\setcctype{by-nc-nd}
\acmJournal{TOG}
\acmYear{2026} \acmVolume{45} \acmNumber{4} \acmArticle{97}
\acmMonth{7} \acmDOI{10.1145/3811378}

% ArXiv
\settopmatter{printacmref=false}
\setcopyright{none}
\renewcommand\footnotetextcopyrightpermission[1]{}

\begin{CCSXML}
<ccs2012>
   <concept>
       <concept_id>10010147.10010178.10010213</concept_id>
       <concept_desc>Computing methodologies~Control methods</concept_desc>
       <concept_significance>500</concept_significance>
       </concept>
   <concept>
       <concept_id>10002950.10003714.10003716</concept_id>
       <concept_desc>Mathematics of computing~Mathematical optimization</concept_desc>
       <concept_significance>300</concept_significance>
       </concept>
   <concept>
       <concept_id>10010147.10010257.10010258.10010261</concept_id>
       <concept_desc>Computing methodologies~Reinforcement learning</concept_desc>
       <concept_significance>500</concept_significance>
       </concept>
   <concept>
       <concept_id>10010147.10010371.10010352</concept_id>
       <concept_desc>Computing methodologies~Animation</concept_desc>
       <concept_significance>500</concept_significance>
       </concept>
 </ccs2012>
\end{CCSXML}

\ccsdesc[500]{Computing methodologies~Control methods}
\ccsdesc[300]{Mathematics of computing~Mathematical optimization}
\ccsdesc[500]{Computing methodologies~Reinforcement learning}
\ccsdesc[500]{Computing methodologies~Animation}

%\keywords{robotic character control, motion tracking, physics-based characters, robotics}

\begin{teaserfigure}
    \centering
    \includegraphics[width=\linewidth]{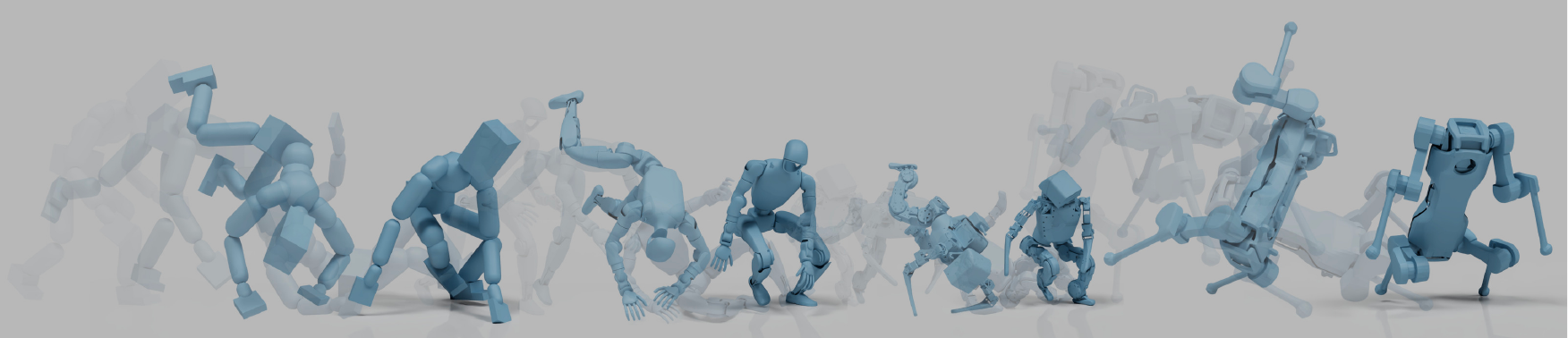}
    \caption{Physics-aware retargeting of human motion (left) onto two humanoid robots (middle) and a quadruped (right) with varying degrees of freedom and vastly different shapes, sizes, and proportions.}
    \Description{.}
  \label{fig:teaser}
\end{teaserfigure}

\maketitle

\input{body-paper}

\end{document}

%% file: body-paper.tex
\section{Introduction}

% Context
Motion data has become a cornerstone of modern animation and robotics, often serving as reference trajectories for imitation learning with deep reinforcement learning (RL)~\cite{peng_deepmimic_2018}. In practice, such reference motions are typically obtained from human motion capture~\cite{mahmood_amass_2019,harvey_robust_2020} or reconstructed from video~\cite{leonardis_tram_2025,goel_humans_2023}. For character control, however, these motions must be adapted to the target embodiment, which can differ substantially in kinematic structure, body shape, mass distribution, and actuation mechanisms.

To bridge the embodiment gap, reference motions are retargeted to characters or robots via a preprocessing step. Optimization-based approaches minimize pose discrepancies between source and target motions~\cite{araujo_retargeting_2025,yang_omniretarget_2025, grandia_doc_2023}.  
However, these methods often require a predefined contact pattern, are prone to local minima, and require substantial manual tuning to scale across diverse motion datasets. 
Learning-based methods offer an alternative by learning direct mappings from source to target motions~\cite{villegas_neural_2018,aberman_skeleton-aware_2020}. However, they often require large datasets of source-target pairs and have primarily been applied to characters with idealized spherical joints, avoiding the complexities of physical characters. 
Additionally, both approaches can produce physically-implausible motions with artifacts like foot sliding, self-penetration, and abrupt joint movements. These artifacts act as a primary source of performance degradation in downstream tasks such as RL policy training~\cite{araujo_retargeting_2025}.

% Our Solution
We instead frame motion retargeting as a reinforcement learning problem within a physics simulation, using a bilevel optimization framework with an RL controller at the lower level, while solving for retargeting parameters in the upper level. The user prescribes coarse correspondences through semantic matching of rigid-body pairs, and the system then solves for optimal offsets between the two embodiments. By jointly optimizing the trajectory and the policy, conflicts between the reference motion and the robot’s morphology can be mitigated, thereby minimizing common retargeting artifacts. Our approach inherently respects physical limitations, accounts for discontinuous contact dynamics, and allows the use of non-differentiable objectives. 

It is important to distinguish retargeting from motion imitation. While frameworks such as DeepMimic~\cite{peng_deepmimic_2018} use RL to track a given kinematic reference, our approach addresses the preceding problem of generating a suitable reference to track, bridging the embodiment gap between robot and human source. Unlike motion imitation, we relax strict dynamic requirements by omitting domain randomization and allowing residual force control (RFC)~\cite{yuan_residual_2020} to act on the root, which facilitates training a single policy across diverse motions (e.g. AMASS~\cite{mahmood_amass_2019}).
Despite these relaxed dynamics, the physics simulation prevents non-physical artifacts like foot sliding, abrupt joint movements, and self-penetration, producing high-quality reference data suitable for downstream tasks.

We demonstrate our retargeting on two humanoid characters, including hardware results on one, and a quadruped (\figref{fig:teaser}). We validate our method using kinematic metrics against baseline humanoid retargeting methods, demonstrate its effectiveness for the downstream task of learning tracking controllers, and show its applicability to quadrupeds. We further analyze the impact of the bilevel optimization and evaluate generalization to unseen motion data. 

Succinctly, we contribute:

\begin{itemize}
    \item A physics-aware, RL-based retargeting framework producing artifact-free motions without making assumptions on contact patterns. 
    \item A bilevel optimization framework jointly adapting parameterized reference motions and learning tracking policies. 
    \item A retargeting parameterization requiring only sparse, semantic rigid-body correspondences defined by the user in a nominal configuration.
\end{itemize}

\begin{figure*}[t]
    \centering
        \includegraphics[width=\linewidth]{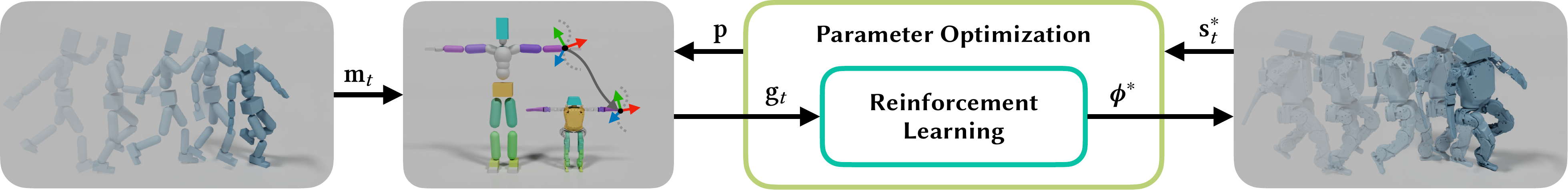}
    \caption{\textbf{Bilevel Optimization for Motion Retargeting.}}
    \label{fig:overview}
\end{figure*}

\section{Related Work}

\paragraph{Motion Retargeting}
Motion retargeting has evolved from kinematic optimization minimizing pose discrepancies~\cite{gleicher_retargetting_1998,schumacher_versatile_2021} to physics-based tracking~\cite{popovic_physically_1999,tak_physically-based_2005,zordan_motion_2002,da_silva_simulation_2008}. Other research has addressed varying proportions~\cite{lyard_motion_2008,liu_surface_2018} and morphologies~\cite{hecker_real-time_2008,chen_motion2motion_2025} via muscle-based models~\cite{ryu_functionality-driven_2021} or interaction-preserving meshes~\cite{ho_spatial_2010,yang_physics-driven_2025,yang_omniretarget_2025}. Data-driven approaches leverage paired supervision~\cite{lee_same_2023,chen_implicit_2025,kim_humanconquad_2022}, semantic labels~\cite{gat_anytop_2025,hu_pose-aware_2024}, or adversarial objectives~\cite{villegas_neural_2018,li_ace_2023,zhu_mocanet_2022,lim_pmnet_2019} to bridge the embodiment gap, often using shared latent spaces~\cite{yan_imitationnet_2023,aberman_skeleton-aware_2020} or geometric refinement~\cite{villegas_contact-aware_2021,zhang_skinned_2023,zhang_skinned_2025,zhang_simulation_2023,reda_physics-based_2023}.

In robotics, retargeting artifacts like foot sliding severely degrade downstream policy training~\cite{araujo_retargeting_2025}. Consequently, specialized methods for humanoids have been developed~\cite{tosun_general_2015,pollard_adapting_2002,darvish_whole-body_2019,ayusawa_motion_2017,rouxel_multicontact_2022}. Additionally, differentiable simulation can be used to optimize for additional effects such as vibration suppression~\cite{hoshyari_vibration-minimizing_2019}. While recent tools, such as PHC~\cite{luo_perpetual_2023}, ProtoMotions~\cite{noauthor_nvlabsprotomotions_2026}, GMR~\cite{araujo_retargeting_2025}, and OmniRetarget~\cite{yang_omniretarget_2025}, streamline sim-to-real transfer, they remain largely kinematic and struggle with temporal coherence. DOC~\cite{grandia_doc_2023} addresses this by optimizing retargeting parameters via differentiable optimal control. Our formulation shares this physics-based focus but differs in three key aspects: we strictly enforce self-collision avoidance, eliminate the need for prescribed contact patterns, and scale to massive datasets using a single retargeting policy.

\paragraph{Physics-based Character Control}
Physics-based control has progressed from trajectory optimization~\cite{hamalainen_online_2015,mordatch_discovery_2012,hodgins_animating_1995,coros_generalized_2010,yin_simbicon_2007} to imitation-based deep reinforcement learning (RL)~\cite{peng_deepmimic_2018}, which is now widely applied in robotics~\cite{grandia_design_2024,fu_humanplus_2024,liao_beyondmimic_2025}. Modern RL methods scale to large datasets~\cite{mahmood_amass_2019,harvey_robust_2020,mason_real-time_2022} and have moved from residual force control~\cite{yuan_residual_2020,luo_dynamics-regulated_2021,zhang_learning_2023} to robust tracking without auxiliary forces~\cite{won_scalable_2020,wang_unicon_2020,fussell_supertrack_2021,serifi_vmp_2024}. However, most frameworks assume morphological equivalence between the source and target characters, with limited exceptions for body shape variation~\cite{won_learning_2019} and terrain-optimized design through grammar-based morphologies~\cite{zhao_robogrammar_2020}. Our method plays a complementary role to these physics-based control strategies by generating the morphologically consistent reference motions they require as input.

\paragraph{Bilevel Optimization}
Bilevel optimization has been applied to various computational design problems that enforce equilibrium constraints for objectives that involve simulation states (see, e.g., ~\cite{Tapia2020,Coros2013,Perez2015,Gjoka2024}). This paradigm has recently gained traction for RL problems to refine latent dynamics~\cite{zhao_bi-level_2024} or optimize reward functions~\cite{xie_kungfubot_2025,lu_deep_2026}. However, we are unaware of the use of stochastic bilevel optimization for retargeting. 

The nested nature of these problems presents significant computational challenges and there exists a wide range of algorithmic approaches~\cite{zhang_introduction_2024}. A standard technique involves calculating the derivative of the lower-level optima using the implicit function theorem. When RL constitutes the lower-level problem, the primary difficulty lies in differentiating the resulting optimal policy or policy rollout with respect to upper-level decision variables. Unlike previous approaches that rely on the implicit function theorem, our work leverages the specific structure of the retargeting problem to derive a simplified gradient estimate.

\section{Bilevel Optimization for Motion Retargeting}
\label{sec:overview}

Our goal is to retarget a dataset of motions from a source morphology to a target robot by simultaneously finding optimal retargeting parameters $\vect{p}$ and learning an optimal retargeting policy $\pi_{\bm{\phi}}$ parameterized by $\bm{\phi}$, for a given $\mathbf{p}$. This can be formulated as the following bilevel optimization problem
\begin{equation}
    \min_{\vect{p} \in \mathcal{P}} \mathcal{L}(\vect{p}, \bm{\phi}^*(\vect{p}) ) \quad \text{subject to} \quad \bm{\phi}^*(\vect{p}) = \arg\max \mathcal{R}(\vect{p}, \bm{\phi}) , 
    \label{eq:bilevel_problem}
\end{equation}
where $\mathcal{P}$ is a convex set to which the parameters are constrained, $\mathcal{L}(\cdot,\cdot)$ is the upper-level loss function, and $\mathcal{R}(\cdot,\cdot)$ is the lower-level reward function.

As illustrated in \figref{fig:overview}, the upper level transforms the source reference motion $\vect{m}_t$ (e.g., human motion capture data) into a parameterized reference motion $\vect{g}_t$ via a mapping governed by the parameters $\vect{p}$. To keep user input to a minimum, we only require semantic correspondences between a sparse set of rigid-body pairs on source and target embodiments. To this end, users select matching rigid bodies on both rigs. The system then automatically solves for the optimal parameters to align the two embodiments. 

At the lower level, we employ RL to train an optimal policy tracking the parameterized reference motion. Rolling out the optimal policy results in a state sequence denoted by $\vect{s}^*_t$. The upper level compares this simulated state with the reference motion and updates the parameters to minimize the error $\ell$, as defined by the upper-level loss function
\begin{equation}
    \mathcal{L}(\vect{p}, \bm{\phi}^*(\vect{p}) ) = \mathbb{E}_{\pi_{\bm{\phi}^*}, \mathbf{s}_0, \mathbf{m}_t} \left[ \ell(\vect{g}_t - \vect{s}^*_t) \right],
\end{equation}
where the expectation is taken over stochastic rollouts of the optimized policy, given the initial states $\mathbf{s}_0$ sampled as described in \secref{sec:rl_for_retargeting}, and motions sampled from the dataset. We omit the explicit dependence of $\pi_{\bm{\phi}^*}$, $\vect{g}_t$, $\vect{s}_0$, and $\vect{s}^*_t$ on $\vect{p}$ to simplify the notation. 

In the following sections, we first present the optimization algorithm used to solve \eqnref{eq:bilevel_problem} (\secref{sec:method_bilevel}), detail the retargeting parameterization (\secref{sec:retargeting}), and describe the RL setup (\secref{sec:rl_for_retargeting}).

\begin{figure*}[t]
    \centering
        \includegraphics[width=\linewidth]{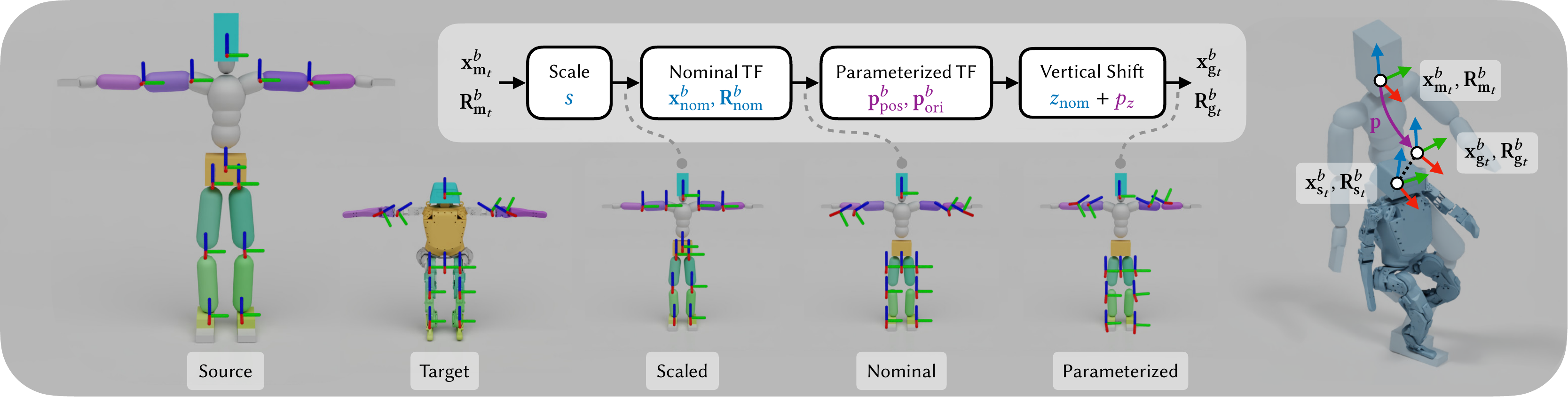}
    \caption{\textbf{Retargeting Parameterization.} A user provides the source and target morphologies in a nominal configuration and defines corresponding rigid-body pairs. A global scale and nominal transformations (TFs) are automatically extracted from the input such that the frames in nominal coordinates align with the corresponding target frames. After this nominal calibration, we introduce parameters in nominal coordinates and a parameterized vertical shift for fine-tuning of source frames during optimization.}
    \label{fig:retargeting_overview}
\end{figure*}

\section{Upper-Level Optimization}
\label{sec:method_bilevel}

A core challenge in our setup is that waiting for the lower-level RL problem to converge before updating the parameters is impractical. We therefore adopt a \textit{single loop} bilevel optimization algorithm \cite{zhang_introduction_2024}, which simultaneously updates the lower- and upper-level decision variables. Specifically, we follow the Two-Timescale Approximation (TTSA)~\cite{mingyi_hong_two-timescale_2023}, and update the upper-level decision variables at each iteration of the RL algorithm according to 
\begin{equation}
\vect{p} \leftarrow P_{\mathcal{P}} \left(\vect{p} - \eta \, \tilde{\text{d}}_{\vect{p}}{\mathcal{L}}  \right), 
\label{eq:p_update}
\end{equation}
where $P_{\mathcal{P}} \left(\cdot\right)$ is the Euclidean projection onto the convex set $\mathcal{P}$, $\eta$ is the step size, and $\tilde{\text{d}}_{\vect{p}}{\mathcal{L}}$ is a gradient estimate of the upper-level loss, approximating the total derivative $\text{d}_{\vect{p}}$ with respect to the parameters. 
In TTSA, this gradient estimate is constructed in several steps. First, as is standard in bilevel optimization \cite{zhang_introduction_2024}, the implicit function theorem is used to derive an expression for $\text{d}_{\vect{p}}{\mathcal{L}}$ based on the optimality conditions of the lower level. Second, since the lower level converges only in the limit, the derived expression is evaluated at the current $\bm{\phi}$ instead of the optimum. Finally, given the stochastic setting, $\tilde{\text{d}}_{\vect{p}}{\mathcal{L}}$ is computed as an estimate using the sampled data available at the current iteration. 

In this work, however, instead of using the implicit function theorem, which requires computing the inverse Hessian of the lower-level problem, we use the structure of the problem to derive a simplified estimate of the upper-level gradient. Consider the total derivative of our error terms
\begin{equation}
    \text{d}_{\vect{p}} \ell(\vect{g}_t - \vect{s}^*_t) = \partial_{\vect{g}_t} \ell \, \, \text{d}_{\vect{p}} \vect{g}_t + \partial_{\vect{s}^*_t} \ell \, \, \text{d}_{\vect{p}} \vect{s}^*_t,
    \label{eq:upper_full_gradient}
\end{equation}
where the sensitivity of the optimal state trajectories $\vect{s}^*_t$ with respect to $\vect{p}$ is challenging to obtain. We avoid computing this sensitivity by making two assumptions: First, we restrict ourselves to loss functions that depend strictly on the difference between $\vect{g}_t$ and $\vect{s}^*_t$, implying the property $\partial_{\vect{s}^*_t} \ell = -\partial_{\vect{g}_t} \ell$ \footnote{To see that this identity holds, we introduce the error, $\boldsymbol{\varepsilon} = \vect{g} - \vect{s}$, omitting super and subscripts, and form the two derivatives, $\partial_{\vect{g}} \ell(\boldsymbol{\varepsilon}) = \partial_{\boldsymbol{\varepsilon}} \ell(\boldsymbol{\varepsilon}) \, \, \partial_{\vect{\mathbf{g}}} \boldsymbol{\varepsilon}$, and, $\partial_{\vect{s}} \ell(\boldsymbol{\varepsilon}) = \partial_{\boldsymbol{\varepsilon}} \ell(\boldsymbol{\varepsilon}) \, \, \partial_{\vect{\mathbf{s}}} \boldsymbol{\varepsilon}$, with $\partial_{\vect{\mathbf{g}}} \boldsymbol{\varepsilon} = \mathbf{I}$ and $\partial_{\vect{\mathbf{s}}} \boldsymbol{\varepsilon} = -\mathbf{I}$.}. Second, given that the optimal RL solution depends on $\vect{p}$ only through $\vect{g}_t$, we can write
\begin{equation}
    \text{d}_{\mathbf{p}} \vect{s}^*_t = \partial_{\mathbf{g}_t} \vect{s}^*_t \, \, \text{d}_{\mathbf{p}} \vect{g}_t,
\end{equation}
where $\partial_{\mathbf{g}_t} \vect{s}^*_t$ represents the change in optimal trajectories given a change in reference motions. We assume this sensitivity takes the form $\alpha \vect{I}$, for some $\alpha \in [0, 1]$, intuitively stating that the resulting optimal trajectories adapt (partially) to changes in reference motions.

Substituting these assumptions into \eqnref{eq:upper_full_gradient} yields a computationally tractable estimate of the upper-level objective gradient
\begin{equation}
    \tilde{\text{d}}_{\mathbf{p}} \ell(\vect{g}_t - \vect{s}^*_t) = (1-\alpha) \, \, \partial_{\mathbf{g}_t} \ell \, \, \text{d}_{\mathbf{p}} \mathbf{g}_t,
    \label{eq:simple_upper_gradient}
\end{equation}
which eliminates the complex sensitivities of the RL solution and is simply a scaled version of the first term in \eqnref{eq:upper_full_gradient}.

We then proceed similarly to TTSA by evaluating \eqnref{eq:simple_upper_gradient} using the current rather than the optimal trajectories and data sampled at the current iteration. Concretely, given a batch $\mathcal{D}$ of state–reference pairs collected from rollouts of the current policy, we compute
\begin{equation}
    \tilde{\text{d}}_{\vect{p}}{\mathcal{L}}  = \frac{1}{|\mathcal{D}|} \sum_{(\vect{s}_{t}, \vect{g}_t) \in \mathcal{D}} \tilde{\text{d}}_{\mathbf{p}} \ell(\vect{g}_t - \vect{s}_t).
\end{equation}

\section{Retargeting Parameterization}
\label{sec:retargeting}

To define the retargeting objective, the user provides the source and target morphologies in a nominal configuration (e.g., a T-pose) and specifies semantic source-target pairs $b$ of rigid bodies (\figref{fig:retargeting_overview}, Source, Target). We assume that the selected pairs are sparse, meaning that not every body on the source has a corresponding body on the target, and vice versa, as is the case for significantly different morphologies. Furthermore, paired bodies do not need to share the same number of adjacent joints. The user also explicitly selects a \textit{root} pair of bodies, relevant for policy training and simulation (see Sec.~\ref{sec:rl_for_retargeting}).

To make our retargeting agnostic to the input, we do not make any assumptions about the location of local coordinate frames on the source and corresponding target body. For source rigs, they usually coincide with the joints, but for robots, their location is less standardized. Our goal is now to map the global position $\vect{x}^b_{\vect{m}_t}$, orientation $\vect{R}^b_{\vect{m}_t}$, linear velocity $\vect{v}^b_{\vect{m}_t}$, and angular velocity $\bm{\omega}^b_{\vect{m}_t}$ of the source frame to quantities that we can compare to the corresponding quantities of the moving target body. 

To define this mapping, we assume the two nominal configurations to be coarsely aligned and apply a global scaling $s$ to the source configuration, which we derive from the root height ratio $s=h_{\text{target}}/h_{\text{source}}$ (\figref{fig:retargeting_overview}, Scaled). In this aligned nominal configuration, we compute the nominal transformation by expressing the relative offset from the scaled source to the target in the source's local coordinate frame
\begin{align}
    \textcolor{nomBlue}{\vect{x}^b_{\text{nom}}} &= (\vect{R}^b_{\text{source}})^T (\vect{x}^b_{\text{target}} - \textcolor{nomBlue}{s} \, \vect{x}^b_{\text{source}}), \\
    \textcolor{nomBlue}{\vect{R}^b_{\text{nom}}} &= (\vect{R}^b_{\text{source}})^T \, \vect{R}^b_{\text{target}},
\end{align}
where $(\vect{x}^b, \vect{R}^b)$ denote the global body frames in the nominal configuration. Note how the frames in nominal coordinates match the ones on the target character after these first two steps (\figref{fig:retargeting_overview}, Nominal, Target). 

To enable our outer-level optimization to make adjustments to the location and orientation of these frames, we introduce position and orientation parameters, $\mathbf{p}^b_{\text{pos}}$ and $\mathbf{p}^b_{\text{ori}}$, in local nominal coordinates. Since reference motions from datasets like AMASS often contain floating or penetration artifacts, we precompute a per-motion nominal vertical offset, $z_{\text{nom}}$, following prior work~\cite{luo_perpetual_2023}. As shown in the supplemental video material, residual floating and penetration remain in the source motions, so we further introduce a learnable per-motion offset $p_z$ to correct artifacts caused by noisy contacts. The full parameterized mapping is therefore
\begin{align}
    \vect{x}^b_{\mathbf{g}_t} &=  \vect{R}^b_{\mathbf{m}_t} (\textcolor{nomBlue}{\vect{R}^b_{\text{nom}}} \textcolor{paramPurple}{\vect{p}^b_{\text{pos}}} + \textcolor{nomBlue}{\vect{x}^b_{\text{nom}}}) + \textcolor{nomBlue}{s}\,\vect{x}^b_{\mathbf{m}_t} + (\textcolor{nomBlue}{z_{\text{nom}}} + \textcolor{paramPurple}{p_z})\vect{e}_z, \\
    \vect{R}^b_{\vect{g}_t} &= \vect{R}^b_{\vect{m}_t} \textcolor{nomBlue}{\vect{R}^b_{\text{nom}}}\text{Exp}(\textcolor{paramPurple}{\vect{p}^b_{\text{ori}}} ), \\
    \vect{v}^b_{\vect{g}_t} &= \bm{\omega}^b_{\vect{m}_t} \times \vect{R}^b_{\vect{m}_t} (\textcolor{nomBlue}{\vect{R}^b_{\text{nom}}} \textcolor{paramPurple}{\vect{p}^b_{\text{pos}}} + \textcolor{nomBlue}{\vect{x}^b_{\text{nom}}} ) + \textcolor{nomBlue}{s}\,\vect{v}^b_{\vect{m}_t}, \\
    \bm{\omega}^b_{\vect{g}_t} &= \bm{\omega}^b_{\vect{m}_t},
\label{eq:ref_motion}
\end{align}
where we highlight \textcolor{nomBlue}{constants that we extract from the nominal configurations} and \textcolor{paramPurple}{parameters that we optimize}. Vector $\vect{e}_z$ is the global unit z-axis, and $\text{Exp}(\cdot)$ is the exponential map, mapping the 3D rotation vector $\vect{p}^b_{\text{ori}}$ to a rotation matrix~\cite{sola_micro_2018}. While our method is agnostic to the specific parameterization, hence interfaces with user-defined variants, we observe that the above parameterized mapping provides a good balance between simplicity and generalization across diverse morphologies and motions.

The convex set $\mathcal{P}$ is defined by constraining the norm of the optimization parameters
\begin{equation}
    \lVert \vect{p}^b_{\text{pos}} \rVert_2 \le \delta_{\text{pos}}, \quad
    \lVert \vect{p}^b_{\text{ori}} \rVert_2 \le \delta_{\text{ori}}, \quad
    | p_{z} | \le \delta_{z}, 
\end{equation}
where $\delta_{\text{pos}}$, $\delta_{\text{ori}}$, and $\delta_{z}$ are the allowed deviations.

With $\vect{g}_t$ fully defined, we can compute differences between the target and simulated state $\mathbf{s}_t$. For position, linear velocity, and angular velocity, we use squared norm loss terms
\begin{equation}
    \ell^b_{\vect{x}} = \lVert \vect{x}^b_{\vect{g}_t} - \vect{x}^b_{\vect{s}_t} \rVert_2^2, \quad \ell^b_{\vect{v}} = \lVert \vect{v}^b_{\vect{g}_t} - \vect{v}^b_{\vect{s}_t} \rVert_2^2, \quad
    \ell^b_{\bm{\omega}} = \lVert \bm{\omega}^b_{\vect{g}_t} - \bm{\omega}^b_{\vect{s}_t} \rVert_2^2.
\end{equation}
With the rotation term, we penalize the geodesic difference between the two rotations \footnote{Even though the loss term is no longer strictly a function of the difference $\vect{g} - \vect{s}$, \eqnref{eq:simple_upper_gradient} can also be derived on the manifold of rotations~\cite{sola_micro_2018}}
\begin{equation}
    \ell^b_{\vect{R}} = \lVert \text{Log} ( (\vect{R}^b_{\vect{s}_t})^T \vect{R}^b_{\vect{g}_t} ) \rVert^2_2,
\end{equation}
where $\text{Log}(\cdot)$ maps a rotation matrix to a 3D rotation vector~\cite{sola_micro_2018}. 

Robots frequently have fewer degrees of freedom than digital characters (e.g., 2 DoF quadruped hip joint). In such cases one might want to ignore rotation errors about unactuated axes. We can achieve this by decomposing orientation error into ``swing'' and ``twist'' components, where ``twist'' is the rotation about a user-specified local axis \cite{dobrowolski_swing_2015}. The orientation loss may then be evaluated on the swing or twist component instead of the full orientation error.

With all tracking losses defined, the upper-level loss function is the sum of all loss terms for all source-target pairs 
\begin{equation}
\ell(\vect{g}_t - \vect{s}_t)
=
\sum_b ( 
w_{\vect{x}} \ell^b_{\vect{x}} +
w_{\vect{R}} \ell^b_{\vect{R}} +
w_{\vect{v}} \ell^b_{\vect{v}} + 
w_{\bm{\omega}} \ell^b_{\bm{\omega}}  ),
\end{equation}
where $w_{\vect{x}}$, $w_{\vect{R}}$, $w_{\vect{v}}$, and $w_{\bm{\omega}}$ are user-specified weights to trade off the relative importance of the error terms. We use the same loss terms in our motion tracking rewards for policy training (see \tabref{tab:rewards}).

\section{Lower-Level Reinforcement Learning}
\label{sec:rl_for_retargeting}

As introduced in \secref{sec:overview}, the lower level of our bilevel optimization trains a policy $\pi_{\bm{\phi}}(\vect{a}_t \,|\, \vect{o}_t, \vect{g}_t)$ to track the parameterized reference motion $\vect{g}_t$. In this section, we will define our actions $\mathbf{a}_t$ and observations $\mathbf{o}_t$, and provide a detailed description of the RL problem. 

\paragraph{Action Space} The policy outputs joint position setpoints $\vect{a}^\text{jts}_t$ for Proportional-Derivative (PD) controllers, and auxiliary wrenches $\vect{w}^{\text{rt}}_t$, consisting of forces $\vect{f}^{\text{rt}}_t$ and torques $\bm{\tau}^{\text{rt}}_t$, at \SI{50}{\hertz}
\begin{equation}
\vect{a}_t \coloneq ( \vect{a}^\text{jts}_t, \vect{w}^{\text{rt}}_t  ) \quad \quad \text{with} \quad \quad \vect{w}^{\text{rt}}_t \coloneq ( \vect{f}^{\text{rt}}_t, \bm{\tau}^{\text{rt}}_t ).
\end{equation}

\noindent The additional wrench, which acts directly on the character's root \cite{yuan_residual_2020}, enables the policy to generalize across large datasets like AMASS~\cite{mahmood_amass_2019}, which contain challenging motions such as handstands that are otherwise infeasible due to morphological differences (e.g., characters without hands). To encourage physical realism, we penalize the usage of this external wrench in the reward function. Additionally, we apply a continuous deadband to the wrench action
\begin{equation}
   \vect{w}^{\text{rt}}_t \coloneq {\operatorname{sgn}( \vect{w}^{\text{rt}}_t) \odot \max\left(0, \operatorname{abs}( \vect{w}^{\text{rt}}_t) - d \right),}
\end{equation}
with threshold $d$, to make it easier for the policy to predict exactly zero forces and torques when residuals are unnecessary. The $\operatorname{sgn}$, $\operatorname{abs}$, and $\operatorname{max}$ functions return the sign, absolute value, or maximum value of each vector component, and the $\odot$ operator multiplies them component-wise.

\paragraph{Proprioceptive State}
The character's proprioceptive state
\begin{equation}
    \vect{o}_t \coloneq ( h^{\text{rt}}_t, \bm{\uptheta}^{\text{rt}}_t, \vect{v}^\text{rt}_t, \bm{\upomega}^\text{rt}_t, \vect{q}_t, \dot{\vect{q}}_t, \vect{a}_{t-1}, \vect{a}_{t-2},\psi_t),
\end{equation}
contains the height $h^{\text{rt}}_t$, the projected gravity vector  $\bm{\uptheta}^{\text{rt}}_t$, and the linear and angular velocities $\vect{v}^\text{rt}_t$ and $\bm{\upomega}^\text{rt}_t$, all extracted from the simulation state $\mathbf{s}_t$ of the robot's root body. The observations also include the joint positions $\vect{q}_t$ and their velocities $\dot{\vect{q}}_t$, and the actions from the previous two time steps, $\vect{a}_{t-1}$ and $\vect{a}_{t-2}$. Additionally, we introduce a \emph{retargeting phase} variable $\psi_t$, whose role we will define below.

\paragraph{Initialization}
\label{sec:rl_initialization}
State-of-the-art motion tracking methods typically rely on Reference State Initialization (RSI)~\cite{peng_deepmimic_2018}, initializing the robot directly to matching root and joint configurations from the reference trajectory. In our setting, however, the source and target morphologies differ, and the initial joint configuration cannot be extracted directly from the reference. Instead of resolving this mismatch with inverse kinematics, we directly learn the initialization with RL. To this end, we set the root state of the robot to the root state of the source character and sample joint positions from a Gaussian distribution around the nominal robot configuration. To let the policy learn to reach the reference pose from this randomized initial configuration, we use a \emph{retargeting phase} variable $\psi_t \in [0, 1]$ that linearly increases from $0$ to $1$ at the start of each episode. During this phase, the reference motion is paused and the policy moves the robot toward the start pose before retargeting begins.
We also use $\psi_t$ for reward blending and data filtering: Our rigid-body tracking rewards are scaled by $\psi_t$ to avoid large penalties while the character prepares for retargeting, as are the penalties on auxiliary forces and torques to allow the character to use this wrench during initialization. Trajectory segments with $\psi_t < 1$ are excluded from the upper-level data batch $\mathcal{D}$ to ensure that retargeting parameters $\vect{p}$ are only optimized once proper tracking is active.

\paragraph{Adaptive Motion Sampling}
We use an adaptive sampling strategy for the motion clips in the dataset, as they vary in difficulty, to prioritize clips where the policy struggles. For each clip, we maintain a failure count based on early episode terminations, which are triggered when the torso position or orientation reward falls below a threshold. During training, clips are sampled with a probability proportional to their failure rate.

\paragraph{Reward Design}

The total discounted reward $\mathcal{R}$ is composed of two terms, a \emph{tracking} reward and a \emph{regularization} reward
\begin{equation}
r_t = r^{\text{tracking}}_t + r^{\text{regularization}}_t.
\label{eq:reward}
\end{equation}
The tracking reward sums up all terms for the rigid-body pairs $b$, with the root pair treated separately (see ~\tabref{tab:rewards}). Following common practice in RL, we add regularization rewards to penalize excessive joint torques and encourage smooth joint actions, helping to avoid vibrations and unnecessary effort. We also penalize the use of the auxiliary wrench on the robot's root, encouraging physical plausibility.  

\input{tables/tab_rewards}

\section{Results}

We evaluate our method on several robotic characters. We compare against two state-of-the-art human motion retargeting methods, GMR~\cite{araujo_retargeting_2025} and OmniRetarget~\cite{yang_omniretarget_2025}. We also present ablation studies of our method, demonstrate retargeting of human data onto a quadruped, and showcase real-world use cases: interactive animation of a physics-based character, and physical robot control.

\paragraph{Implementation Details.}

\begin{figure}[t]
    \centering
    \includegraphics[width=\linewidth,trim={0 360 320 0}, clip]{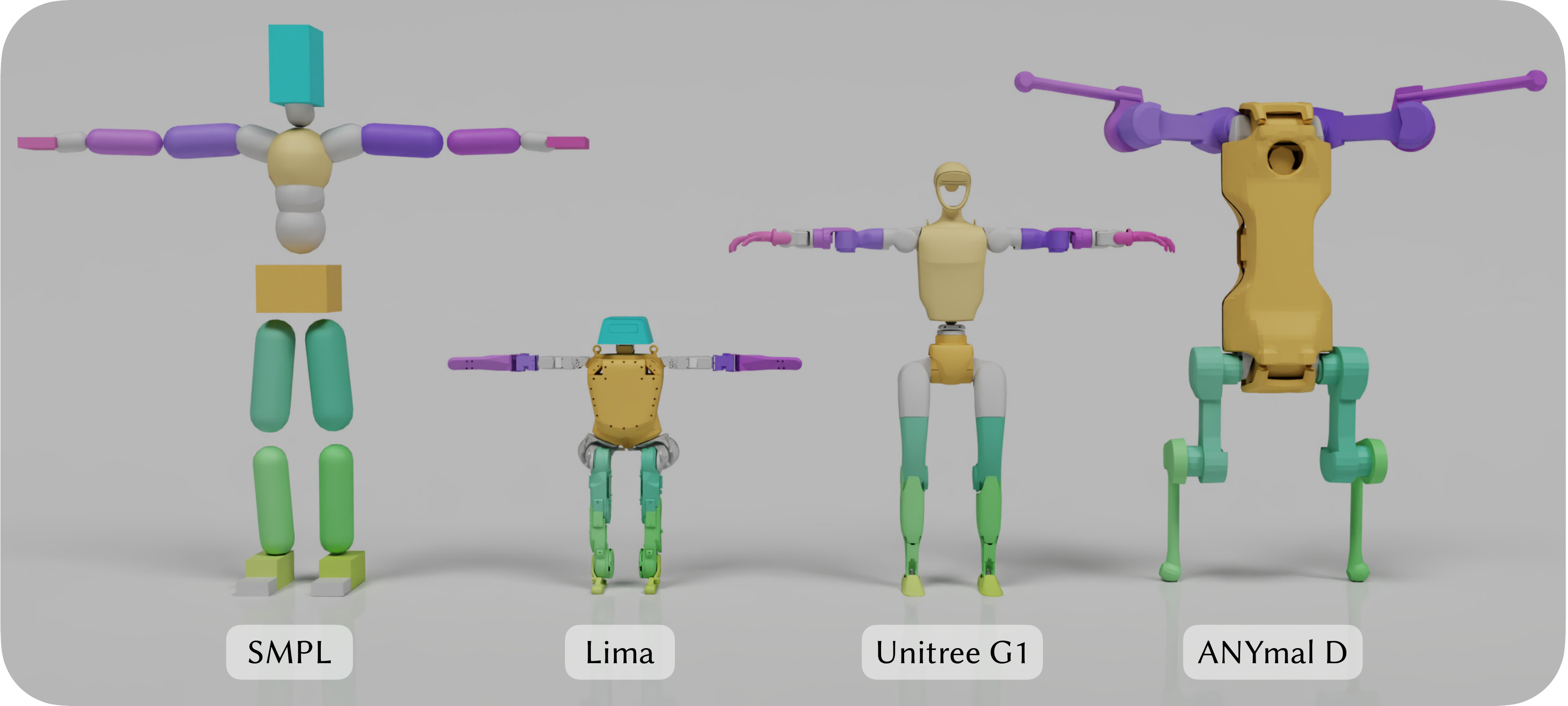}
    \caption{\textbf{Semantic correspondences and nominal configurations.} for the SMPL body model and our target robots: Unitree G1, Lima, and ANYmal D.}
    \label{fig:correspondences}
\end{figure}

We retarget motions onto two humanoids of different scale: \textit{Unitree G1} (1.27 m, 35 kg, 29 DoF), and \textit{Lima} (0.84 m, 16.2 kg, 20 DoF), a custom small-scale robot. For the Unitree G1, we apply the baseline methods using their provided hyperparameter sets. For Lima, we use the frames after the nominal alignment step as input to the baseline methods. The kinematic correspondences between the source motions and the robot bodies are visualized in \figref{fig:correspondences}. While all our robotic targets have fewer DoF than the human source, the formulation also applies when the target has more DoF, as RL regularization (acceleration, torque, action rate) ensures well-behaved solutions even in under-constrained settings, where additional reward terms could help adjust the results towards a preferred aesthetic goal. Wherever robots have fewer DoF than the source character, we use the twist-swing decomposition. The correspondences were assigned based on structural similarity, without iterative refinement.
Unless stated otherwise, we use the AMASS dataset~\cite{mahmood_amass_2019}. Adopting the filtering criteria from PHC~\cite{luo_perpetual_2023}, we remove sequences with human-object interactions or excessive noise. The resulting curated dataset is used directly without additional preprocessing.

We train our policies using PPO~\cite{schulman_proximal_2017} with an adaptive learning rate~\cite{rudin_learning_2022}. Both the policy and value function are modeled using multi-layer perceptron (MLP) networks with ELU activations, consisting of three layers with $512$ units each. 
%Observations are normalized using a running mean, following the standard practice in PPO~\cite{schulman_proximal_2017}. 
All simulations are performed using Isaac Sim, running $4,096$ environment instances in parallel on a single RTX $5090$ GPU. We run our method for $20$k iterations (${\sim}$\SI{6}{\hour}). Hyperparameters are listed in \tabref{tab:hyperparams}.

\input{tables/tab_hyperparams}

\subsection{Baseline Comparison}

We compare our method to two state-of-the-art human motion retargeting methods, GMR~\cite{araujo_retargeting_2025} and OmniRetarget~\cite{yang_omniretarget_2025}, on two humanoid robots at different scales.

\paragraph{Kinematic Evaluation}
\label{sec:kinematic_evaluation}

\begin{figure}[t]
    \centering
        \includegraphics[width=\linewidth]{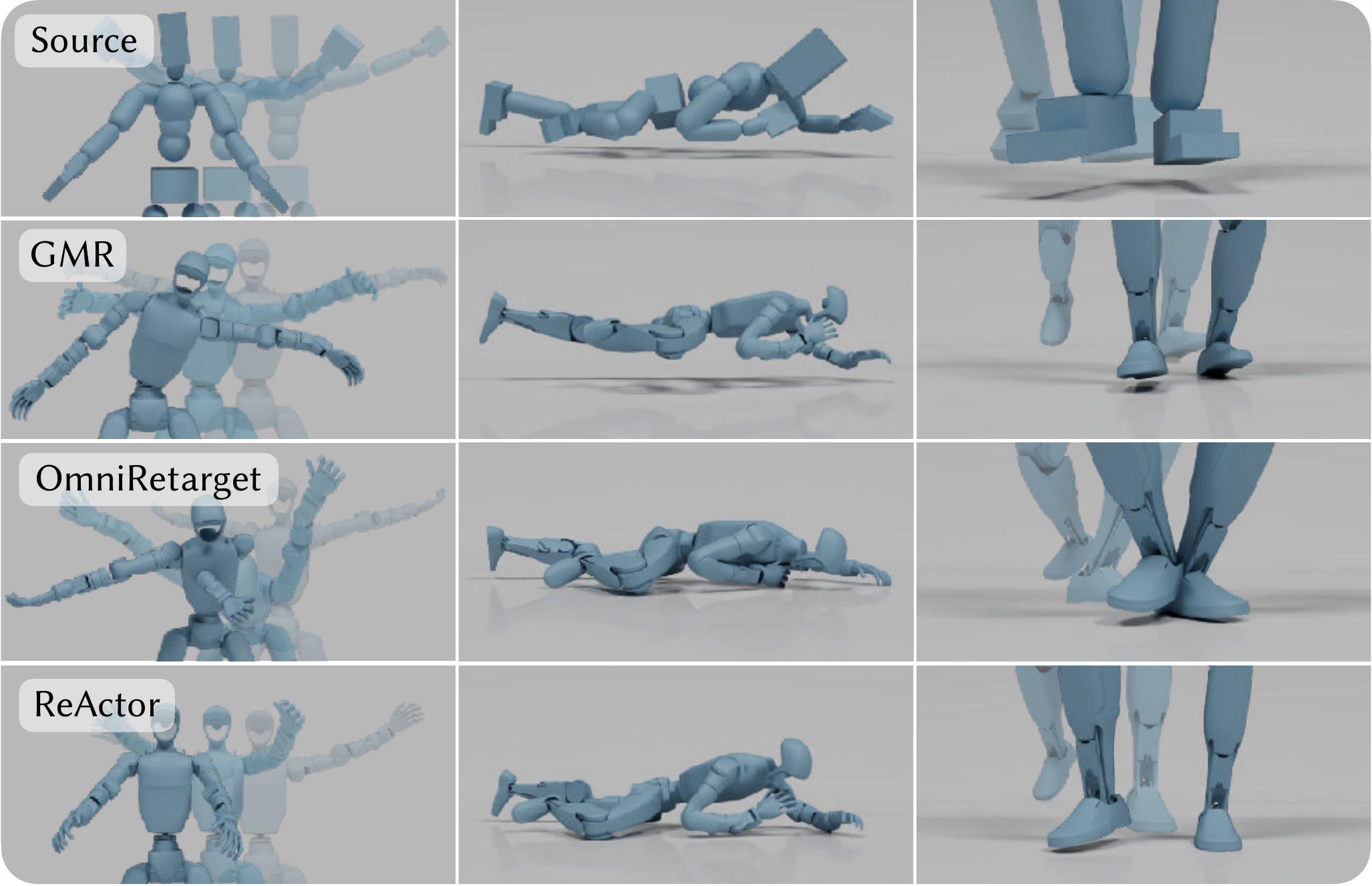}
    \caption{\textbf{Kinematic Artifacts.} Local solver minima, often present near kinematic singularities or joint limits (left). Floating (middle) and self-penetration (right) artifacts.}
    \label{fig:kinematic_artifacts}
\end{figure}

First, we compare our method against the baselines through several kinematic metrics, as detailed in \tabref{tab:kin_eval_metrics}. 
\begin{table}
    \centering{}
    \caption{\textbf{Kinematic Evaluation Metrics.} Based on OmniRetarget, with \textit{self-penetration} and \textit{foot floating} added. Where reference contact state is used, this is estimated following \cite{shimada_physcap_2020}.}
    \label{tab:kin_eval_metrics}
    \footnotesize
    \begin{tabular}{l p{5.5cm}}
    \toprule
    \textbf{Metric} & \textbf{Description} \\
    \midrule
    \textit{Ground Penetration} & Fraction of motion frames where penetration exceeds \SI{0.01}{\meter}. Reported penetration depth is mean across violating frames. If multiple simultaneous ground contacts, record maximum penetration depth per frame. \\
    \textit{Self-Penetration} & Time, depth computed as for ground-penetration. Collisions within same kinematic chain ignored, to remove false positives. \\
    \textit{Foot Sliding} & Mean linear velocity of robot's feet during reference ground contact phases. \\
    \textit{Foot Floating} & Mean of minimum distances between robot's foot and ground during reference ground contact. \\
    \bottomrule
    \end{tabular}
\end{table}
The quantitative results are summarized in~\tabref{tab:retargeting_eval}, with best- and worst-case variability reported in the supplemental material. Additionally, common baseline artifacts are visualized in~\figref{fig:kinematic_artifacts} and the supplemental video. Our method significantly outperforms both baselines across all metrics on both humanoid platforms. The most severe failures of the optimization-based baselines occur when the solver converges to local minima, typically near kinematic singularities and joint limits. As illustrated in the first column of~\figref{fig:kinematic_artifacts}, the arm over-rotates near a singular configuration, reaches a joint limit, and gets stuck in a local minimum. This leads to severe artifacts and self-penetration, making the retargeted motion unusable.

Both baselines exhibit ground contact artifacts. GMR relies purely on data preprocessing and does not apply any height correction during retargeting. As a result, it exhibits both ground penetration and floating (\figref{fig:kinematic_artifacts}, second row). OmniRetarget prevents ground penetration by strictly correcting the motion height at each frame while enforcing ground contact to match heuristically-estimated contact patterns from the reference motion. However, these objectives may conflict, leading to foot floating, as seen  in~\tabref{tab:retargeting_eval}. In contrast, our method avoids both artifacts by design. Note that non-zero values for floating and sliding persist, as our method does not strictly enforce contact matching with the reference.
Our method avoids self-penetration by explicitly accounting for contact dynamics within the simulation during retargeting, unlike the baselines. 

As seen in~\tabref{tab:retargeting_eval}, OmniRetarget struggles with the Lima platform, likely due to its non-uniform scaling (approximately half human height but similar width) and non-standard root alignment, which degrades contact estimation. Although per-motion tuning could mitigate these issues, OmniRetarget fails to generalize across the full dataset when using nominal reference parameters. However, GMR behaves worse for G1 but performs better for Lima. In terms of computational cost, parallel training and retargeting on the AMASS dataset requires ${\sim}$\SI{6.5}{\hour} on a single GPU, which is comparable to running OmniRetarget (${\sim}$\SI{7}{\hour}) and GMR (${\sim}$\SI{5}{\hour}) on a CPU.

\input{tables/tab_retargeting}

\paragraph{Downstream RL Performance}
\label{sec:downstream_rl}

A central observation from prior work~\cite{araujo_retargeting_2025} is that the kinematic quality of retargeted motions strongly influences the success of downstream Reinforcement Learning (RL) training. 
We train RL tracking policies on the retargeted data from each method and use identical hyperparameters without any method-specific tuning. Unlike \cite{yang_omniretarget_2025}, which evaluates on a selected subset of 39 sequences from AMASS \cite{mahmood_amass_2019}, we train and evaluate on the entire filtered AMASS dataset \cite{luo_perpetual_2023}. We refer to our supplemental material for details about these tracking policies. 
We report the success rate, measured by the ability of the policy to complete the motion without triggering the termination criteria used during RL training \cite{yang_omniretarget_2025}, in \tabref{tab:retargeting_eval}. 
Additionally, we report root mean squared errors for the root position, root orientation, and joint position tracking.
For each motion, we initialize the episodes with random starting frames and run for 5 seconds, unless the episode ends early due to the termination criteria. Note that the success rate of our approach outperforms the baselines on both G1 and Lima. Similarly, the tracking policies show smaller joint position and root pose errors when trained with data from \OURMETHOD.

\subsection{Ablation Studies}

\begin{figure}[t] 
    \centering
        \includegraphics[width=\linewidth]{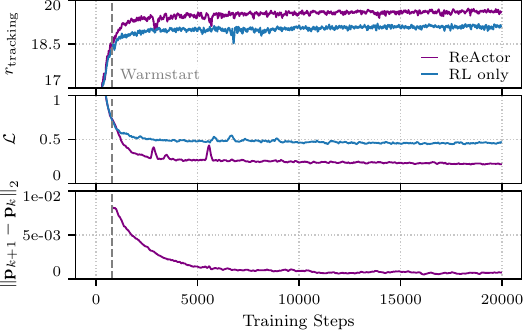}
        \caption{\textbf{Training Curves with and without the Bilevel Optimization.} Tracking reward, upper-level loss, and parameter update rate during training with and without bilevel optimization. The update rate shows outer-loop convergence and is zero without bilevel optimization, as the parameters remain static.}
    \label{fig:bilevel_abalation}
\end{figure}

\begin{figure}[t] 
    \centering
        \includegraphics[width=\linewidth]{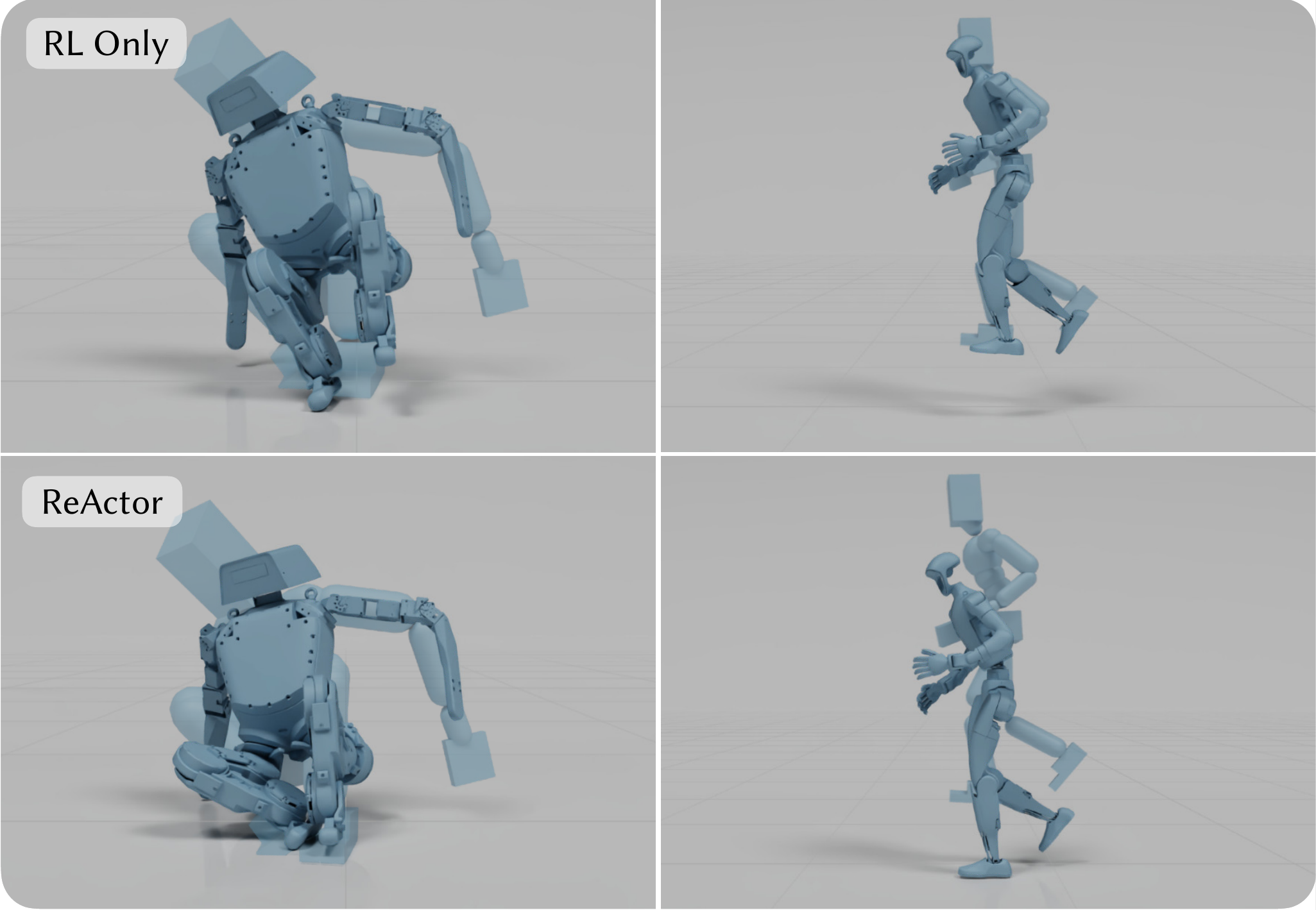}
        \caption{\textbf{Qualitative Comparison With and Without Bilevel Optimization.} With bi-level optimization, results have fewer retargeting artifacts.}
        \label{fig:bilevel_visual}
\end{figure}

\paragraph{Parameterization} We study the impact of the proposed parameterization by training policies with and without the upper-level optimization, as shown in~\figref{fig:bilevel_abalation},~\figref{fig:bilevel_visual}, and the video. The bilevel optimization consistently reduces tracking loss and leads to higher tracking rewards, confirming the effectiveness of the proposed parameterization. Qualitative results further demonstrate that the robot tracks motions in a more natural and physically consistent manner, as the policy and reference are iteratively refined to better align with each other.

In the video, we further explore the impact of source-robot correspondences and target parameterization. We replace the dense set of correspondences with a sparse set of only the root and end effectors. The result deviates further from the source, as would be expected, but the method remains stable. As an additional robustness test against mapping mismatches in the input, we assign the left robot hand to track the motion of the head of the source, and the method remains stable.

We also compare our parameterization against an orientation-only baseline. While both produce stable results, the inclusion of positional targets is beneficial for motions like jumping, where it improves the timing and fidelity of lift-off and touchdown.

\paragraph{Generalization} We evaluate how well a policy, trained on the full AMASS dataset, generalizes to unseen motion data, enabling a single retargeting policy to be reused across datasets and within real-time user applications. As there is no absolute retargeting ground truth, we train a retargeting policy directly on the test data and use its output as a pseudo-ground-truth reference.
To this end, we randomly partition the 100STYLE dataset by selecting 50\% of the motions for training and reserving the remaining half for testing. We train separate retargeting policies for Lima on the 100STYLE training subset, the full AMASS dataset, and the 100STYLE test subset (pseudo-ground-truth). We compare the retargeting errors between these policies on the pseudo-ground-truth reference in \tabref{tab:generalization}. Note that these errors are smaller than in \tabref{tab:retargeting_eval}, as they evaluate the retargeting policy itself, which uses residual forces. In contrast, \tabref{tab:retargeting_eval} evaluates a separate downstream tracking policy trained without residual forces.

\paragraph{External Force}

\begin{figure}[t]
    \centering
    \includegraphics[width=\linewidth]{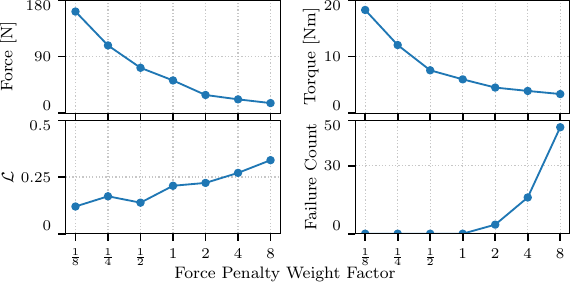}
    \caption{\textbf{Effect of the External Force Penalty.} Mean over motions of the maximum applied forces and torques, the mean upper-level loss $\mathcal{L}$, and the total number of failures as a function of the force penalty weight.}
    \label{fig:force_weight}
\end{figure}

The external force penalty weight is the most sensitive hyperparameter. Other parameters, such as regularization weights, primarily suppress high-frequency jitter without significantly affecting motion quality. Thus, we analyze the force penalty weight trade-off on retargeting performance in \figref{fig:force_weight}. Increasing the penalty encourages greater physical realism but can lead to failures on more challenging motions, whereas weaker penalties improve retargeting success at the cost of physical plausibility. We report the mean over motions of the maximum applied forces and torques, together with the mean upper-level loss $\mathcal{L}$ and the total failure count. A motion is considered a failure if, after training, the retargeting policy triggers the termination condition on the root pose (orientation error > \SI{45}{\degree} $\vee$ position error > \SI{1}{\meter}).

\input{tables/tab_generalization}

\subsection{Use Cases}

\paragraph{Retarget Human Data onto a Quadruped}
We demonstrate the versatility of our approach by applying our method to a quadruped, \textit{ANYmal D} (50 kg, 12 DoF). See \figref{fig:correspondences} for semantic correspondences, and the video for retargeted motions. Even with widely different embodiments, the motion's visual appearance is preserved, while also providing valuable insights into the method's limitations. As the embodiment gap becomes larger, more nuanced reward tuning becomes necessary. In the video, we show an example where ANYmal gives up tracking in favor of reducing external force usage.

\paragraph{Interactive Animation of Physics-based Character}
To demonstrate practical applicability, we deploy the retargeting policy in an interactive user setting, where an artist modifies a motion sequence on the fly while the policy retargets the motion in real time to the robot morphology (see video). The system runs at $\SI{88.3}{\hertz}$, exceeding real-time requirements and enabling seamless, responsive motion editing. We also see applications in the real-time retargeting of a performance onto a robot during a capture session.

\paragraph{Physical Robot Control}
As shown in~\tabref{tab:retargeting_eval}, our method significantly improves downstream reinforcement learning performance. We further validate this by deploying goal-conditioned tracking policies, trained with a DeepMimic-style reward formulation~\cite{peng_deepmimic_2018} on data retargeted by \textbf{\OURMETHOD}, directly on the physical Lima robot. We evaluate a range of challenging motions within the robot’s hardware limits, showcased in the video. The successful sim-to-real transfer demonstrates that our retargeting pipeline produces motion references suitable for real-world robotic deployment.

\section{Conclusion}

This paper introduces a novel bilevel optimization framework that effectively bridges the embodiment gap between human motion and diverse robotic morphologies. By framing retargeting as a joint problem where retargeting parameters and an RL tracking policy are optimized simultaneously, the system significantly reduces common artifacts. This integrated approach, supported by a simplified gradient estimate for computational efficiency, allows the framework to produce physically plausible motions that serve as high-quality reference data for downstream imitation learning tasks. Moreover, we also see applications in training generative motion models and real-time retargeting, e.g. during live mocap sessions. 

The current external force penalty weight serves as a tuning parameter that enables a user to prioritize either physical realism or successful retargeting of the most extreme motions in the dataset. Ultimately, the retargeting of physically impossible motions remains an ill-posed problem, where it is unclear if the robot should walk up the virtual staircase or if the retargeting method should project the motion to the ground. Regardless, providing more user-control over the result is desirable.

While ReActor establishes a strong foundation for physics-aware retargeting, several avenues for future research remain. Currently, the optimized parameterization is assumed to be constant over time. Exploring time-varying parameterizations could further increase the solution space, though it may introduce new challenges. Moreover, automating the semantic correspondence selection could further reduce user input. More broadly, the bilevel formulation presented herein holds significant promise for more complex scenarios where reference tracking must be balanced with auxiliary objectives, such as obstacle avoidance, manipulation, or even automated robot design. We believe that such tasks should be approached in an integrated, bilevel manner, rather than as sequences of decoupled steps.

% Bibliography
\bibliographystyle{ACM-Reference-Format}
\bibliography{ReActor}

% Appendix
\appendix

\section{Downstream RL Policy Details}

Following prior work, we evaluate the retargeting methods on a downstream tracking task by training RL policies on the retargeted motions, with tracking performance serving as a proxy for motion quality \cite{yang_omniretarget_2025, liao_beyondmimic_2025}. We train the RL policies following the DeepMimic framework \cite{peng_deepmimic_2018}, where the policy is conditioned on a motion reference and optimized using explicit tracking rewards. The reward terms used for training are detailed in \tabref{tab:rewards_rl}.

We measure the success rate based on the training termination criteria, as proposed in \cite{yang_omniretarget_2025}. A trial is considered a failure if the robot’s root deviates by more than \SI{1}{\meter} from the target root position or if the geodesic distance between the current and target root orientation exceeds \SI{45}{\degree}.

\input{tables/tab_rewards_oracle}

\section{Performance Variability}

\tabref{tab:variability} reports best- and worst-case results (min/max of the per-motion mean) for Lima, complementing the mean and standard deviation reported in the main paper.

\input{tables/tab_variability}

%% file: tables/tab_rewards.tex
\begin{table}[tb]
\begin{center}
    \caption{\textbf{Weighted Reward Terms.} %$\vect{v}_t$ is the linear velocity of the root, and 
    $\bm{\uptau}^{\text{jts}}_t$ and $\ddot{\vect{q}}_t$ are joint torques and accelerations.}
    \label{tab:rewards}
    \footnotesize
    \begin{tabular}{l | l | l}\toprule 
    \textbf{Name}                  &  \textbf{Reward Term} & \textbf{Weight}\\
    \midrule
    \midrule
    \multicolumn{3}{c}{\textit{Motion Tracking}}  \\
    \midrule
    Root position xy & %$\exp(-200.0 \cdot \lVert \vect{x}_{\text{root}} - \hat{\vect{x}} _{\text{root}} \rVert_2^{2})$ 
    $-\ell^\text{rt}_{x,y}$ & $2.0$ \\
    Root height & %$\exp(-200.0 \cdot \lVert \vect{x}_{\text{root}} - \hat{\vect{x}} _{\text{root}} \rVert_2^{2})$ 
    $-\ell^\text{rt}_{z}$ & $10.0$ \\
    Root orientation & %$-\lVert {\vect{R}}_{\text{root}} - \hat{\vect{R}}_{\text{root}} \rVert_2^{2}$ 
    $-\ell^\text{rt}_{\vect{R}}$ & $2.0$ \\
    Root lin vel. & %$-\lVert {\vect{v}}_{\text{root}} - \hat{\vect{v}}_{\text{root}}  \rVert_2^{2}$ 
    $-\ell^\text{rt}_{\vect{v}}$ & $0.5$ \\
    Root ang. vel. & %-\lVert {\bm{\upomega}}_{\text{root}} - \hat{\bm{\upomega}}_{\text{root}} \rVert_2^{2}
    $-\ell^\text{rt}_{\bm{\omega}}$ & $0.5$ \\
    Rbs position &  %$-\phi \cdot\lVert \vect{x}_{\text{rbs}} - \hat{\vect{x}}_{\text{rbs}} \rVert_2 ^2 $ 
     $- \ell^b_{\vect{x}}$ & $ 2.0 \cdot \psi_t $ \\
    Rbs orientation & % $-\phi \cdot r_{\text{rot}} $ 
    $-\ell^\text{b}_{\vect{R}}$ & $ 2.0 \cdot  \psi_t $ \\
    Survival  &  $1.0$ & $20$ \\
    \midrule
    \multicolumn{3}{c}{\textit{Regularization}}  \\
    \midrule
    Joint torques  &  $- \lVert \bm{\uptau}^{\text{jts}}_t \rVert_2 ^2 $ & $1.0 \cdot 10^{-4}$ \\
    Joint acc.     &  $- \lVert \ddot{\vect{q}}_t \rVert_2 ^2 $ & $1.0 \cdot 10^{-6}$ \\
    Joint action rate    &  $- \lVert \vect{a}^{\text{jts}}_t - \vect{a}^{\text{jts}}_{t-1} \rVert_2 ^2 $ & $ 1.0 \cdot 10^{-2} $ \\
    Joint action acc.    &  $- \lVert \vect{a}^{\text{jts}}_t - 2\vect{a}^{\text{jts}}_{t-1} + \vect{a}^{\text{jts}}_{t-2} \rVert_2 ^2 $ & $ 1.0 \cdot 10^{-2} $ \\
    Root Force & $- \lVert \vect{f}^{\text{rt}}_t \rVert_1$& $\psi_t \cdot 10^{-2}$\\
    Root Torque & $- \lVert \bm{\tau}^{\text{rt}}_t \rVert_1$ & $\psi_t \cdot 10^{-2}$\\
    \bottomrule
    \end{tabular}
\end{center}
\end{table}

%% file: tables/tab_hyperparams.tex
\begin{table}[tb]
\begin{center}
\caption{\textbf{Hyperparameters.} The PPO hyperparameters and bilevel optimization parameters used to train the tracking policy.}
\label{tab:hyperparams}
\footnotesize
\begin{tabular}{ll|ll}
\toprule
\textbf{Param.} & \textbf{Value} & \textbf{Param.} & \textbf{Value} \\ \midrule \midrule
Num. iterations & $20\,000$ & 
$\delta_{\text{pos}}$ & $0.5$ \\
Batch size $(\text{envs.}\times\text{steps})$ & $4096\times24$ & 
$\delta_{\text{ori}}$ & $0.5$ \\
Num. mini-batches & $4$ & 
$\delta_{\text{z}}$ & $0.5$ \\
Num. epochs & $5$ & 
$w_{\mathbf{x}}$ & $10.0$ \\ 
Clip range & $0.2$ & 
$w_{\mathbf{R}}$ & $1.0$ \\  
Entropy coefficient & $0.0025$ & 
$w_{\bm{\omega}}$, $w_{\mathbf{v}}$ & $0.0$ \\
Discount factor & $0.97$ &
$d$ & 0.1 \\
GAE discount factor & $0.95$ & & \\
Desired KL-divergence & $0.009$ & & \\
Max gradient norm & $1.0$ & & \\ \bottomrule
\end{tabular}
\end{center}
\end{table}

%% file: tables/tab_retargeting.tex
\begin{table*}[t]
    \centering
    \caption{\textbf{Retargeting Evaluation.} Comparison against baselines on the PHC-filtered AMASS subset. We report mean and standard deviation for ground penetration, self-penetration, foot sliding, and foot floating. For the downstream RL task, we report success rate, and root position, root orientation, and joint root mean squared tracking errors.}
    \label{tab:retargeting_eval}
    \footnotesize
    \setlength{\tabcolsep}{4pt} 
    \begin{tabular}{l cc cc c c cccc}
        \toprule
        & \multicolumn{2}{c}{\textbf{Ground-Pen.}} & \multicolumn{2}{c}{\textbf{Self-Pen.} } & \textbf{Foot Slide} & \textbf{Foot Float}& \multicolumn{4}{c}{\textbf{Downstream RL}} \\
        \cmidrule(lr){2-3} \cmidrule(lr){4-5} \cmidrule(lr){6-6} \cmidrule(lr){7-7} \cmidrule(lr){8-11}
        \textbf{Method} & Time $\downarrow$ & Depth [cm] $\downarrow$ & Time $\downarrow$ & Depth [cm] $\downarrow$ & Vel. [cm/s] $\downarrow$ & Height [cm] $\downarrow$ & Success [\%] $\uparrow$ & Pos. [cm] $\downarrow$ & Ori. [deg] $\downarrow$ & Joints [deg] $\downarrow$\\ 
        \midrule
        
        \multicolumn{8}{l}{\textit{Unitree G1}} \\ 
        GMR & $0.53\pm0.30$ & $2.74\pm0.92$ & $0.07\pm0.15$ & $5.63\pm2.80$ & $1.25\pm3.86$ & $0.34\pm0.95$ & $89.93$ & $2.99\pm4.94$ & $4.48\pm5.09$ & $~~9.79\pm12.06$\\
        OmniRetarget & $\mathbf{0.00\pm0.00}$ & $\mathbf{0.00\pm0.00}$ & $0.12\pm0.13$ & $3.27\pm1.55$ & $2.00\pm1.23$ & $0.49\pm0.19$ & $95.51$ & $1.84\pm3.18$ & $3.32\pm2.77$ & $6.62\pm7.17$\\
        \textbf{\OURMETHOD} & $\mathbf{0.00\pm0.00}$ & $\mathbf{0.00\pm0.00}$ & $\mathbf{0.00\pm0.00}$ & $\mathbf{0.00\pm0.00}$ & $\mathbf{0.17\pm1.25}$ & $\mathbf{0.12\pm0.32}$ & $\mathbf{97.45}$ & $\mathbf{1.11\pm2.39}$ & $\mathbf{1.87\pm1.68}$ & $\mathbf{4.22\pm2.22}$\\
        
        \midrule
        \multicolumn{8}{l}{\textit{Lima}} \\ 
        GMR & $0.34\pm0.41$ & $2.42\pm0.43$ & $0.04\pm0.13$ & $3.56\pm1.89$ & $1.97\pm4.42$ & $1.27\pm2.59$ & $91.23$ & $3.53\pm5.25$ & $4.38\pm5.36$ & $10.45\pm18.10$\\
        OmniRetarget & $\mathbf{0.00\pm0.00}$ & $\mathbf{0.00\pm0.00}$ & $0.09\pm0.23$ & $3.89\pm2.17$ & $2.40\pm2.09$ & $0.31\pm0.23$ & $79.85$ & $5.86 \pm 9.15$ & $6.32\pm6.44$ & $14.10\pm16.37$\\
        \textbf{\OURMETHOD} & $\mathbf{0.00\pm0.00}$ & $\mathbf{0.00\pm0.00}$ & $\mathbf{0.00\pm0.00}$ & $\mathbf{0.00\pm0.00}$ & $\mathbf{0.47\pm2.38}$ & $\mathbf{0.02\pm0.08}$ & $\mathbf{95.07}$ & $\mathbf{1.46\pm1.88}$ & $\mathbf{3.00\pm2.07}$ & $\mathbf{4.38\pm2.92}$\\
        \bottomrule
    \end{tabular}
\end{table*}

%% file: tables/tab_generalization.tex
\begin{table}[tb]
\begin{center}
    \caption{\textbf{Generalization Evaluation.} Retargeting policy tracking errors on the 100STYLE test set, measured against a pseudo-ground-truth reference (a policy trained on the test set). Rows show a policy trained on the 100STYLE training split and one trained on AMASS.}
    \footnotesize
    \begin{tabular}{l|ccc}
    \toprule
    Training & Pos. [cm] $\downarrow$ & Ori. [deg] $\downarrow$ & Joints [deg] $\downarrow$ \\
    \midrule
    100STYLE & $0.12 \pm 0.07$ & $5.57 \pm 2.46$ & $5.79 \pm 2.04$ \\
    AMASS    & $0.19 \pm 0.15$ & $6.18 \pm 2.01$ & $6.93 \pm 1.53$ \\
    \bottomrule
    \end{tabular}
    \label{tab:generalization}
\end{center}
\end{table}

%% file: tables/tab_rewards_oracle.tex
\begin{table}[tb]
\begin{center}
    \caption{\textbf{Reward Terms for Downstream RL Training.} The root position is given by $\vect{x}^\text{rt}$ and the root height is $z^\text{rt}$. The root orientation matrix is $\vect{R}^\text{rt}$, the root's linear and angular velocities are $\vect{v}^\text{rt}$ and $\bm{\omega}^\text{rt}$, respectively. We denote rigid body positions as $\vect{x}^b$ and rigid body orientations as $\vect{R}^b$. The terms $\bm{\uptau}^{\text{jts}}_t$ and $\ddot{\vect{q}}_t$ are joint torques and accelerations. The policy actions are given by $\vect{a}^{\text{jts}}_t$. Note that in this case $\vect{g}_t$ the retargeted trajectory and $\vect{s}_t$ denotes the simulation state of the downstream RL policy.}
    \label{tab:rewards_rl}
    \footnotesize
    \begin{tabular}{l | l | l | l}\toprule 
    \textbf{Name} & \textbf{Reward Term} & \textbf{Weight G1} & \textbf{Weight Lima}\\
    \midrule
    \midrule
    \multicolumn{4}{c}{\textit{Motion Tracking}}  \\
    \midrule
    Root position xy & $-\lVert \vect{x}^\text{rt}_{\vect{g}_t} - \vect{x}^\text{rt}_{\vect{s}_t} \rVert_2^2$ & $5.0$ & $5.0$  \\
    Root height & $-( z^\text{rt}_{\vect{g}_t} - z^\text{rt}_{\vect{s}_t} )^2$ & $5.0$ & $5.0$ \\
    Root orientation & $-\lVert \text{Log} ( (\vect{R}^\text{rt}_{\vect{s}_t})^T \vect{R}^\text{rt}_{\vect{g}_t} ) \rVert^2_2$ & $3.0$ & $3.0$  \\
    Root lin vel. & $-\lVert \vect{v}^\text{rt}_{\vect{g}_t} - \vect{v}^\text{rt}_{\vect{s}_t} \rVert_2^2$ & $0.5$ & $0.5$  \\
    Root ang. vel. & $-\lVert \bm{\omega}^\text{rt}_{\vect{g}_t} - \bm{\omega}^\text{rt}_{\vect{s}_t} \rVert_2^2$ & $0.5$ & $0.5$  \\
    Rbs position & $-\lVert \vect{x}^b_{\vect{g}_t} - \vect{x}^b_{\vect{s}_t} \rVert_2^2$ & 5.0 & 5.0  \\
    Rbs orientation & $-\lVert \text{Log} ( (\vect{R}^b_{\vect{s}_t})^T \vect{R}^b_{\vect{g}_t} ) \rVert^2_2$ & 2.5 & 2.5  \\
    Survival  & $1.0$ & 10.0 & 1.0 \\
    \midrule
    \multicolumn{4}{c}{\textit{Regularization}}  \\
    \midrule
    Joint torques  & $- \lVert \bm{\uptau}^{\text{jts}}_t \rVert_2 ^2 $ & $1.0 \cdot 10^{-4}$ & $1.0 \cdot 10^{-3}$ \\
    Joint acc.     & $- \lVert \ddot{\vect{q}}_t \rVert_2 ^2 $ & $2.5 \cdot 10^{-8}$ & $2.5 \cdot 10^{-6}$ \\
    Joint action rate    & $- \lVert \vect{a}^{\text{jts}}_t - \vect{a}^{\text{jts}}_{t-1} \rVert_2 ^2 $ & 0.15 & 3.0 \\
    Joint action acc.    & $- \lVert \vect{a}^{\text{jts}}_t - 2\vect{a}^{\text{jts}}_{t-1} + \vect{a}^{\text{jts}}_{t-2} \rVert_2 ^2 $ & $ 1.0 \cdot 10^{-2} $ & 1.0 \\
    \bottomrule
    \end{tabular}
\end{center}
\end{table}

%% file: tables/tab_variability.tex
\begin{table}[tb]
\begin{center}
    \caption{\textbf{Performance Variability (Lima).} Best and worst case (min/max of the per-motion mean) for the kinematic metrics reported in the main paper.}
    \label{tab:variability}
    \footnotesize
    \begin{tabular}{l ccc}\toprule
    \textbf{Metric} & \textbf{ReActor} & \textbf{OmniRetarget} & \textbf{GMR} \\
    \midrule
    Ground Pen. [cm]  & $\mathbf{0.0~/~0.0}$ & $0.0~/~0.0$   & $0.0~/~9.3$ \\
    Self Pen. [cm]    & $\mathbf{0.0~/~0.0}$ & $0.0~/~17.2$  & $0.0~/~15.3$ \\
    Foot Slide [cm/s] & $\mathbf{0.0~/~52.3}$ & $0.0~/~98.6$  & $0.0~/~95.5$ \\
    Foot Float [cm]   & $\mathbf{0.0~/~9.4}$ & $0.0~/~21.4$  & $0.0~/~32.4$ \\
    \bottomrule
    \end{tabular}
\end{center}
\end{table}